\documentclass[letterpaper]{article} 
\usepackage{aaai25}  
\usepackage{times}  
\usepackage{helvet}  
\usepackage{courier}  
\usepackage[hyphens]{url}  
\usepackage{graphicx} 

\usepackage{natbib}  
\usepackage{caption} 
\usepackage{algorithm}
\usepackage{algorithmic}

\usepackage{newfloat}
\usepackage{listings}
\DeclareCaptionStyle{ruled}{labelfont=normalfont,labelsep=colon,strut=off} 
\floatstyle{ruled}
\newfloat{listing}{tb}{lst}{}
\floatname{listing}{Listing}

\title{Personalizing Exposure Therapy via Reinforcement Learning}
\author {
    Athar Mahmoudi-Nejad\textsuperscript{\rm 1, 2},
    Matthew Guzdial\textsuperscript{\rm 1, 2},
    Pierre Boulanger\textsuperscript{\rm 1}
}
\affiliations {
    \textsuperscript{\rm 1}Computing Science Department, University of Alberta\\
    \textsuperscript{\rm 2}Alberta Machine Intelligence Institute\\
    athar1@ualberta.ca, guzdial@ualberta.ca, pierreb@ualberta.ca
}

\usepackage{bibentry}
\usepackage{booktabs}
\usepackage{xcolor}
\usepackage{colortbl}
\usepackage{xfp}

\usepackage[bottom]{footmisc}

\newcommand{\heatmapcolor}[1]{%
    \ifnum \fpeval{#1<3}=1 \cellcolor{yellow!30}%
    \else \ifnum \fpeval{#1<5}=1 \cellcolor{yellow!60}%
    \else \ifnum \fpeval{#1<7}=1 \cellcolor{orange!70}%
    \else \cellcolor{red!90}%
    \fi\fi\fi
    #1%
}
\begin{document}

\maketitle

\begin{abstract}
Personalized therapy, in which a therapeutic practice is adapted to an individual patient, can lead to improved health outcomes. 
Typically, this is accomplished by relying on a therapist's training and intuition along with feedback from a patient. 
However, this requires the therapist to become an expert on any technological components, such as in the case of Virtual Reality Exposure Therapy (VRET). 
While there exist approaches to automatically adapt therapeutic content to a patient, they generally rely on hand-authored, pre-defined rules, which may not generalize to all individuals.
In this paper, we propose an approach to automatically adapt therapeutic content to patients based on physiological measures. 
We implement our approach in the context of virtual reality arachnophobia exposure therapy, and rely on experience-driven procedural content generation via reinforcement learning (EDPCGRL) to generate virtual spiders to match an individual patient. 
Through a human subject study, we demonstrate that our system significantly outperforms a more common rules-based method, highlighting its potential for enhancing personalized therapeutic interventions.
\end{abstract}

%

\section{Introduction}

Anxiety disorders are the most prevalent mental illness, with an estimated 301 million people living with them in 2019~\cite{yang2021global}.
Anxiety disorders result in notable challenges including decreased well-being, low productivity, reduced quality of life, and frequent medical visits~\cite{ost2023cognitive}.
Cognitive-behavioral therapy, particularly Exposure Therapy (ET), is effective in treating anxiety by gradually exposing individuals to feared situations to reduce sensitivity~\cite{kaczkurkin2022cognitive}.

Virtual Reality Exposure Therapy (VRET) is a modern form of ET that immerses individuals in virtual environments, offering safe and practical exposure to feared stimuli, such as virtual spiders for arachnophobia~\cite{carl2019virtual, maples2017use, valmaggia2016virtual}. VRET effectively induces physiological anxiety responses (e.g.,  blood pressure, muscle vibration, heart rate, and sweating)~\cite{bun2017low, vsalkevicius2019anxiety}.

VRET applications are mainly non-adaptive, wherein all users experience the same content. For example, all users see the same virtual spider for arachnophobia or encounter the same high-standing location for fear of heights. However, personalized treatment has shown greater benefits for individuals~\cite{smits2019personalized}. Adaptive VRET attempts to select specialized content to elicit the user's desired anxiety levels in an individual, offering a more personalized experience. Current adaptive VRET is mostly either therapist-guided~\cite{kampmann2016exposure, bualan2020towards} or rules-based~\cite{bian2019design, repetto2013virtual}, limiting scalability and personalization~\cite{smits2019personalized}.
Therefore, an automatic framework that effectively adapts to individuals to achieve desired anxiety levels without extensive therapist intervention or predefined rules is desirable.

This work introduces an Experience-Driven Procedural Content Generation (EDPCG) framework using Reinforcement Learning (RL) for VRET targeting arachnophobia~\cite{yannakakis2011experience}. The framework aims to automatically generate and adjust virtual spiders to achieve desired anxiety levels without extensive therapist intervention. This approach represents the first PCGRL application in rehabilitation, differing from previous rules-based methods by adapting to individual user responses.
Focusing on arachnophobia, the RL agent dynamically generates spiders to elicit specific anxiety levels defined by therapists. To evaluate the effectiveness of our framework, we conducted a human subject study in which we evaluated our framework in comparison to a rules-based method. 
We found that our framework was effective at achieving desired anxiety levels, indicating that it may be feasible in clinical settings.
However, we only tested the system's ability to reach desired anxiety levels, not its effect on therapeutic outcomes.
Finally, we conducted an experiment to demonstrate the necessity of personalizing spiders to each user.

We present the following contributions:
\begin{itemize}
   \item Developing an end-to-end personalized EDPCGRL framework, which incorporates real-time human interaction to automatically adapt a VR environment to a user.
   \item  Conducting a human subject study to evaluate the effectiveness of our framework compared to a rules-based method by collecting both subjective and objective measures during the experiments.
   \item  Demonstrating that different participants exhibit anxiety responses to different attributes of spiders, emphasizing the necessity for personalized treatment for each participant.  

\end{itemize}

\section{Related Work}

\paragraph{Physiological Measures: }
To assess anxiety levels in real-time, studies have utilized physiological measures such as heart rate variability (HRV)~\cite{cheng2022heart, harrewijn2018heart, alvares2013reduced, chalmers2014anxiety, gaebler2013heart} and electrodermal activity (EDA)~\cite{christian2023electrodermal, kimani2019addressing, kritikos2019anxiety, lonsdorf2017don, vsalkevicius2019anxiety, sevil2017social, nikolic2018bumping, owens2015can}. EDA refers to the autonomic changes in the electrical properties of the skin. It consists of Skin Conductance Response (SCR) and Skin Conductance Level (SCL), both of which correlate with anxiety and emotional responses~\cite{boucsein2012electrodermal}. Studies have shown strong correlations between SCL and self-reported anxiety in VRET~\cite{wilhelm2005mechanisms, muhlberger2008virtual}, and SCR amplitudes have been used to approximate stress and arousal~\cite{suso2019virtual, pick2018autonomic, yee2015insecure, albayrak2023fear, sevincc2018language}. In this paper, we employed SCL to assess participants’ anxiety.

\paragraph{Procedural Content Generation (PCG) for Rehabilitation:}
Procedural Content Generation (PCG) refers to the algorithmic generation of content.
PCG algorithms have been employed in adaptive rehabilitation. Examples include a ski-slalom game for balance improvement~\cite{dimovska2010towards}, a self-adaptive shooter game for amblyopia~\cite{correa2014new}, and adaptive tasks for upper limb rehabilitation~\cite{hocine2015adaptation}. Other applications involve VR labyrinths for emotional self-awareness~\cite{i2018toward} and gait rehabilitation customization~\cite{lyu2023procedural}. 
These studies typically use constructive or rules-based PCG methods~\cite{guzdial2022classical}, relying on predefined rules under the assumption of homogeneity among subjects, i.e., they can predict all subjects’ behaviors or interactions with the system. In contrast, we assume that subjects are unknown and exhibit different behaviors; therefore, the system needs to generate and adapt game content dynamically based on each subject’s needs.
Our approach is based on PCG via reinforcement learning (PCGRL) which leverages reinforcement learning (RL) for dynamic, personalized content generation~\cite{guzdial2022reinforcement, khalifa2020pcgrl}.

\paragraph{Virtual Reality Exposure Therapy (VRET) for Arachnophobia:}
VRET has been effective in reducing fear of spiders and avoidance behaviors~\cite{ cote2009cognitive, michaliszyn2010randomized, minns2019immersive, miloff2019automated,  dyrdal2022virtual}. Early studies showed significant improvements in VRET groups compared to controls~\cite{michaliszyn2010randomized, miloff2019automated}. While most studies pre-design scenarios with attributes like spider size and behavior, \citet{kritikos2021personalized} introduced an adaptive VR environment using rules-based methods to adjust spider attributes based on EDA measures. However, their approach assumed uniform responses, unlike our reinforcement learning-based method that personalizes content dynamically.

\paragraph{Adaptive Virtual Reality (VR) using Physiological Measures:}
Adaptive VR environments using physiological measures have been developed for various health conditions. Examples include a virtual beach adjusting scenarios based on HRV for tension-type headaches~\cite{fominykh2018conceptual}, a driving simulator for autism spectrum disorder modifying difficulty based on engagement~\cite{bian2019design}, and a breathing training system using physiological feedback~\cite{tu2018breathcoach}. Other studies adjusted visual cues in VR for Generalized Anxiety Disorders based on heart rate~\cite{repetto2013virtual, gorini2010virtual}. Unlike these rule-based adaptations, our framework employs reinforcement learning to dynamically tailor the VR environment to individual user responses without prior assumptions.

\section{Proposed Framework}

\begin{figure}[t]
\centering
\includegraphics[width=0.95\columnwidth]{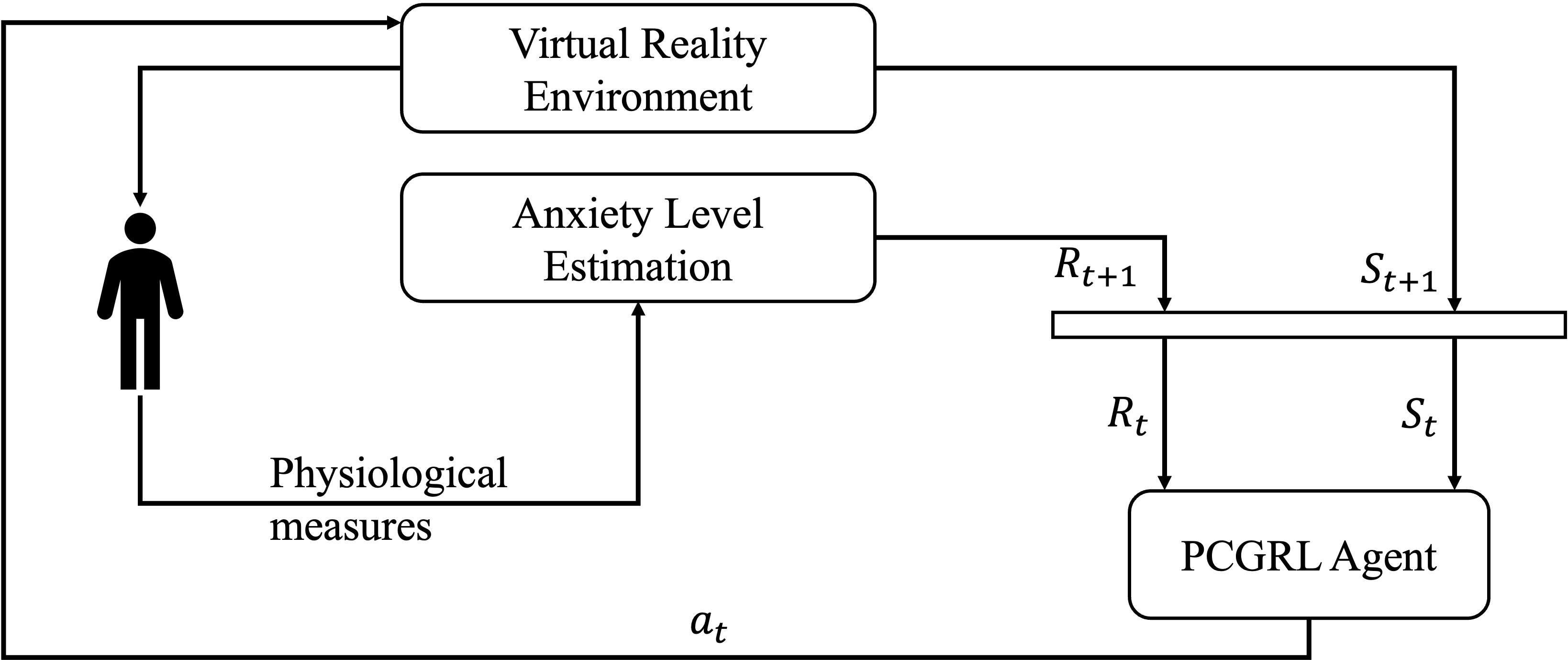} 
\caption{Proposed EDPCGRL framework}
\label{fig: system}
\end{figure}

We introduce the Experience-Driven Procedural Content Generation via Reinforcement Learning (EDPCGRL) framework for Virtual Reality Exposure Therapy (VRET), illustrated in Figure \ref{fig: system}. This framework dynamically adjusts the VR environment in real-time based on the user's anxiety levels using a PCGRL approach.

The VR environment contains adaptive variables relevant to the specific exposure therapy, such as the height for fear of heights. Anxiety estimation approximates the user's current anxiety in real-time and compares it to the desired level set by a therapist, providing a reward signal for the system. The PCGRL component acts as a content generator, modifying the adaptive variables to maximize the reward. 
This framework enables personalized and flexible VRET by automatically tailoring the VR content to each individual's needs without extensive therapist intervention.

\paragraph{Arachnophobia Implementation:}
For arachnophobia, we developed a VR environment featuring a 3D spider model with six adaptive attributes—Locomotion, Amount of Movement, Closeness, Largeness, Hairiness, and Color—based on prior research~\cite{lindner2019so} (see Table~\ref{Table: spider_attr}). Each attribute has 2-3 ordinal values to allow dynamic adjustments. User anxiety is estimated by measuring Skin Conductance Level (SCL) using an Electrodermal Activity (EDA) sensor, which is then discretized and mapped to a scale from 0 to 10. This current anxiety level is compared to a desired level set by a therapist to calculate a reward, structured as a normal distribution scaled between $[-1,1]$, where the desired level yields a reward of 1, decreasing exponentially as deviations occur.

The PCGRL agent functions as an automatic content generator, adjusting the spider's attributes based on the estimated anxiety level. In our RL setup, each state represents a unique combination of the spider’s attributes. Actions are defined as the incremental increase or decrease of a single attribute at a time. The agent employs a tabular Q-learning algorithm with an epsilon-greedy ($\epsilon = 0.05$) action selection policy~\cite{sutton2018reinforcement}. We found this to be sufficient given the relatively small number of states ($|S| = 486$).

\begin{table}[h]
    \centering
    \begin{tabular}{|p{2cm}|p{5.5cm}|}
        \hline
        \textbf{Attribute}       & \textbf{Possible Values}          \\
        \hline
        Locomotion               & Standing (0), Walking (1), Jumping (2)    \\ \hline
        Amount of Movement        & Slightly (0), Moderate (1), Too much (2)       \\ \hline
        Closeness                & Far away (0), In the middle (1), Very close (2)       \\ \hline
        Largeness                & Small (0), Medium (1), Large (2)    \\ \hline
        Hairiness                & Without (0), With (1)       \\ \hline
        Color                    & Grey (0), Red (1), Black (2)       \\ 
        \hline
    \end{tabular}
    \caption{The spider attributes based on~\citet{lindner2019so} and defined possible values.}
    \label{Table: spider_attr}
\end{table}

\section{Human Subject Study}

\begin{figure*}[t]
  \centering
  \begin{minipage}[b]{0.32\textwidth}
    \centering
    \includegraphics[width=\textwidth]{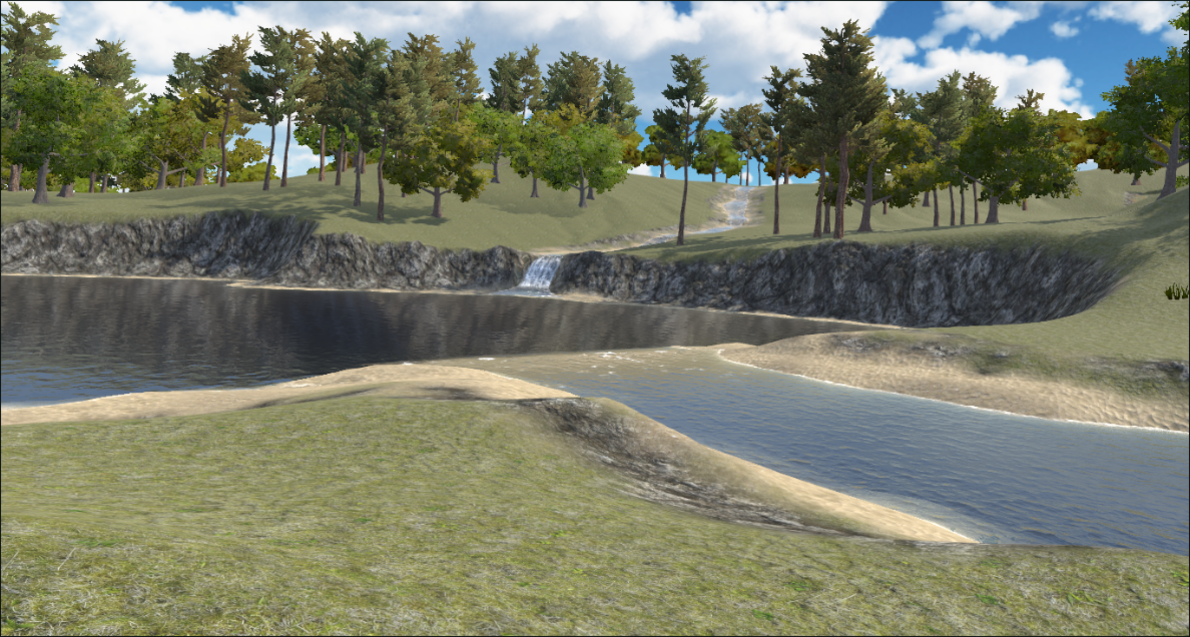}
    \caption*{Relaxing}
    \label{fig: relaxing}
  \end{minipage}
  \hfill
  \begin{minipage}[b]{0.32\textwidth}
    \centering
    \includegraphics[width=\textwidth]{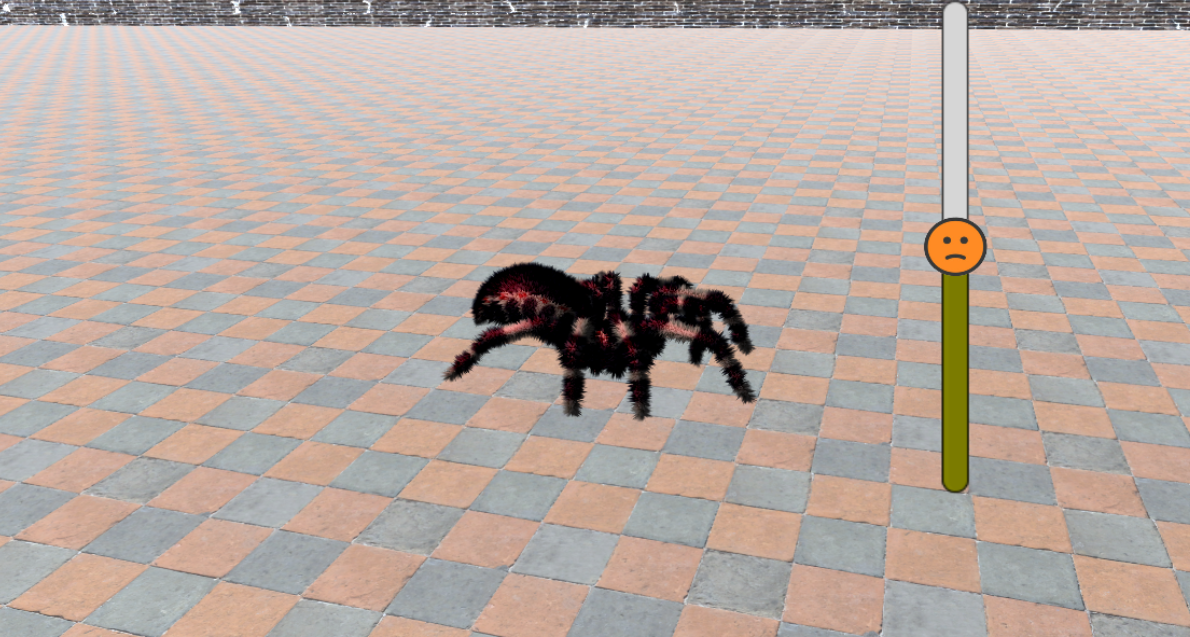}
    \caption*{Anxious}
  \end{minipage}
  \hfill
  \begin{minipage}[b]{0.32\textwidth}
    \centering
    \includegraphics[width=\textwidth]{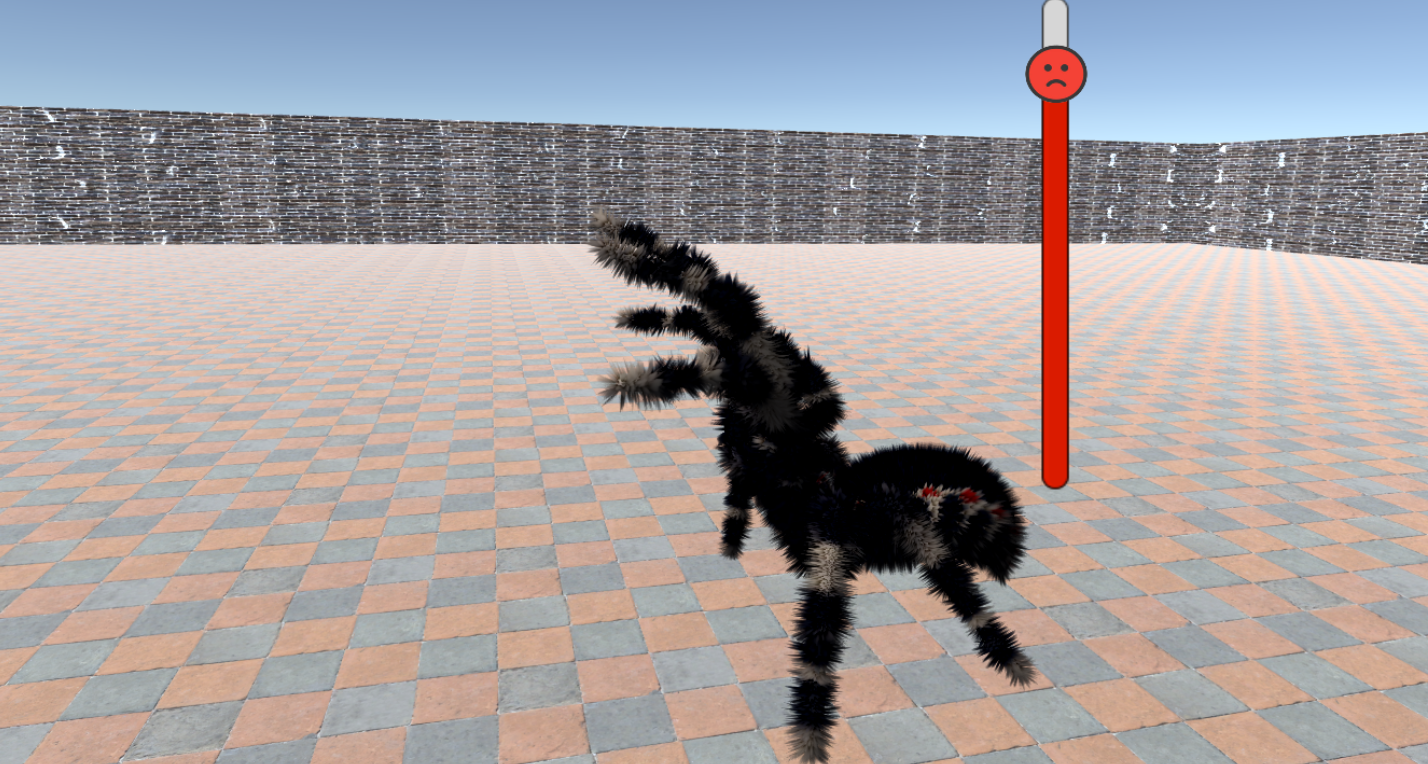}
    \caption*{Anxious}
  \end{minipage}
  \hfill
  \caption{The relaxing and anxious virtual reality environments for our human subject study.}
  \label{fig:VR enivonments}
\end{figure*}

To evaluate the effectiveness of our EDPCGRL framework, we conducted a human subject study focused on arachnophobia, comparing our RL method against a standard rules-based approach. The study aimed to address two primary questions: whether the framework could elicit specific desired anxiety levels in participants and how its performance compared to a rules-based method.

\paragraph{Rules-based Method:}
Rules-based methods are currently the most common Artificial Intelligence approach for adaptive VRET applications~\cite{zahabi2020adaptive}. For our study, we implemented a rules-based method inspired by \citet{kritikos2021personalized}, as it is the only existing research on adaptive VRET for arachnophobia. We adapted their approach, which utilizes a ``correction factor'' representing the divergence between an individual's actual EDA and the desired EDA level. Specifically, we defined the correction factor as the difference between the current anxiety level and the desired anxiety level divided by 10, resulting in a value ranging from -1 to 1. To align with our RL method, we discretized these attributes. Recognizing the potential for this approach to make multiple changes concurrently, we opted to alter only one attribute at a time to avoid any potential risk to participants and to better measure the impact of each change. Therefore, we adapted the rules-based method to randomly select one attribute from those originally chosen for modification.

\paragraph{Study Procedure:}
We began the experiment by exposing participants to a relaxing nature environment for 120 seconds to stabilize their physiological measures (see Figure~\ref{fig:VR enivonments}). They were then exposed to an anxious environment for 280 seconds, featuring a spider whose attributes were dynamically adjusted using either our RL method or the rules-based method (see Figure~\ref{fig:VR enivonments}). After the first anxious session, participants returned to the relaxing environment for another 120 seconds, followed by a second anxious session of 280 seconds employing the alternate adaptive method. Each anxious session included a low-anxiety segment (desired level 3) and a high-anxiety segment (desired level 7). Throughout the sessions, anxiety levels were monitored using Skin Conductance Level (SCL) and self-reported Subjective Units of Distress (SUDs). SUDs~\cite{wolpe1990practice} is a measure of stress and anxiety, utilizing a self-assessment scale ranging from 0 (no distress or total relaxation) to 100 (highest anxiety/distress ever felt). Upon completing the VR sessions, participants filled out a questionnaire to assess their experiences.

\paragraph{Participants:}
The study involved twenty-two non-arachnophobic participants (11 males and 11 females) aged between 18 and 35 years. Participants exhibited varied gaming habits, with two playing daily, eight weekly, four monthly, and eight less frequently. Seventeen participants had prior experience with virtual reality (VR), while five did not. The participants were counterbalanced to experience either the RL method first or the rules-based method first, ensuring an equal distribution and mitigating potential order effects.

\section{Results}

\begin{figure*}[t]
  \centering
  \begin{minipage}[b]{0.49\textwidth}
    \centering
    \includegraphics[width=\textwidth]{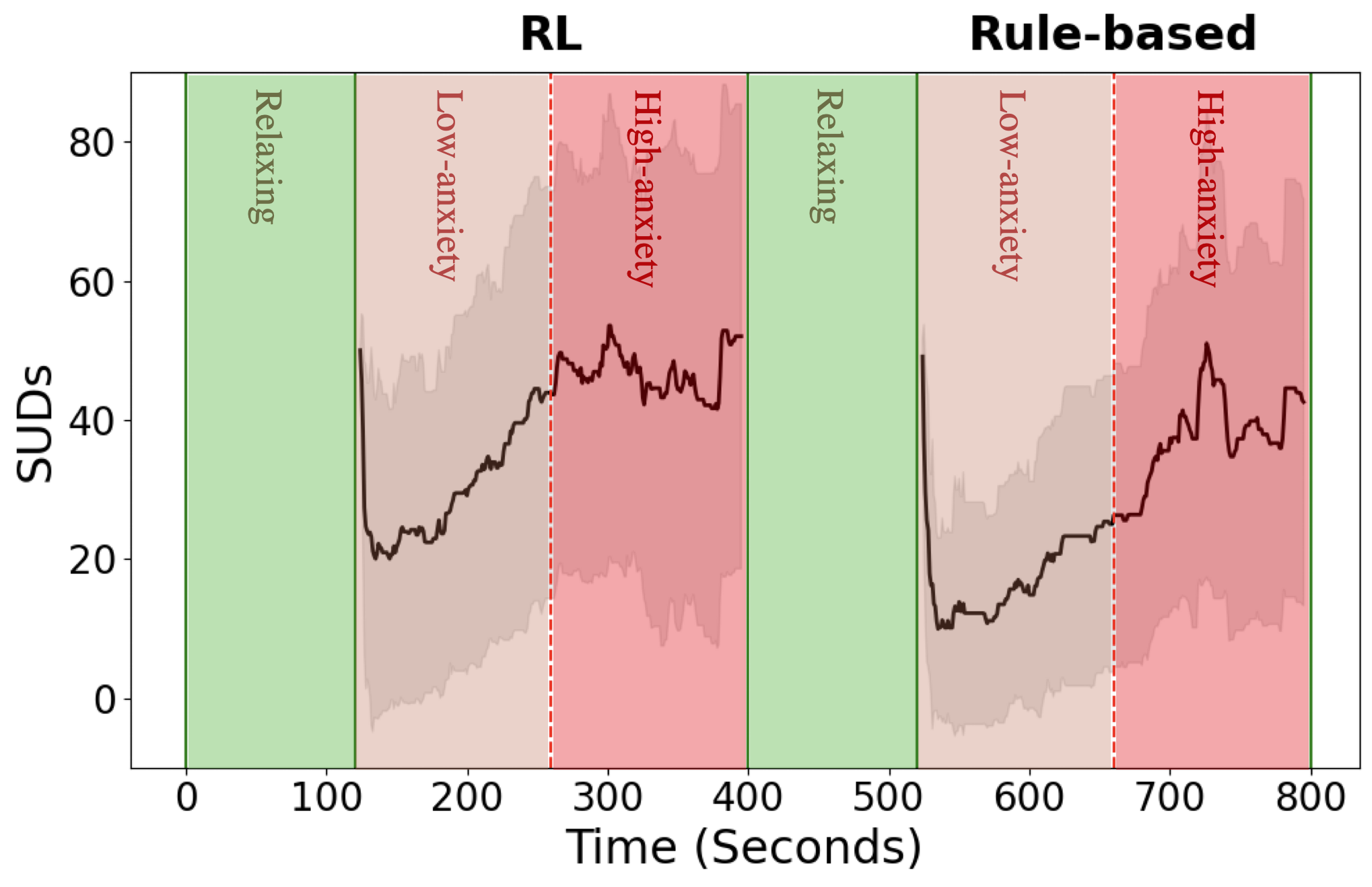}
  \end{minipage}
  \hfill
  \begin{minipage}[b]{0.49\textwidth}
    \centering
    \includegraphics[width=\textwidth]{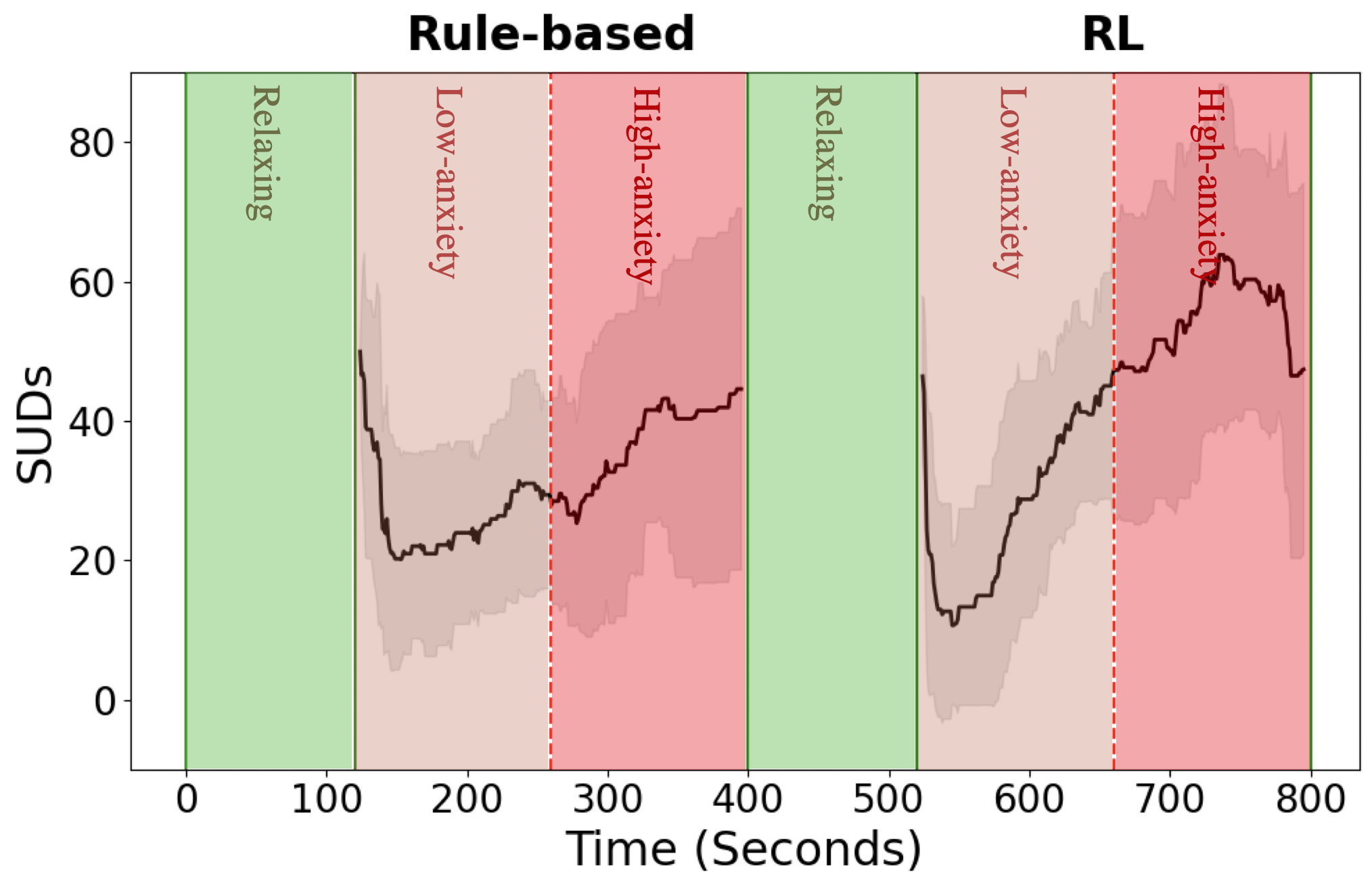}
  \end{minipage}
  \hfill
  \begin{minipage}[b]{0.49\textwidth}
    \centering
    \includegraphics[width=\textwidth]{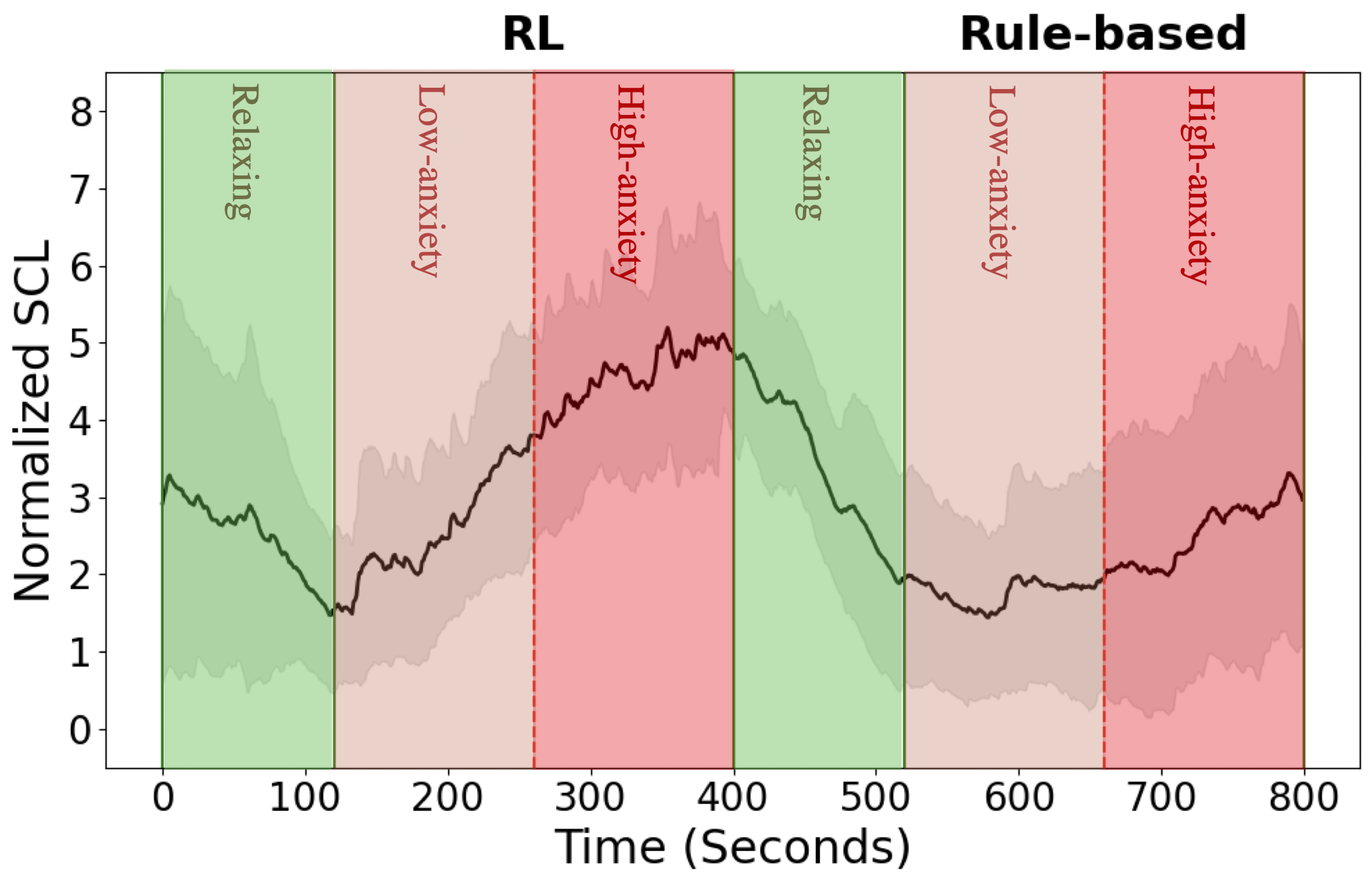}
  \end{minipage}
  \hfill
  \begin{minipage}[b]{0.49\textwidth}
    \centering
    \includegraphics[width=\textwidth]{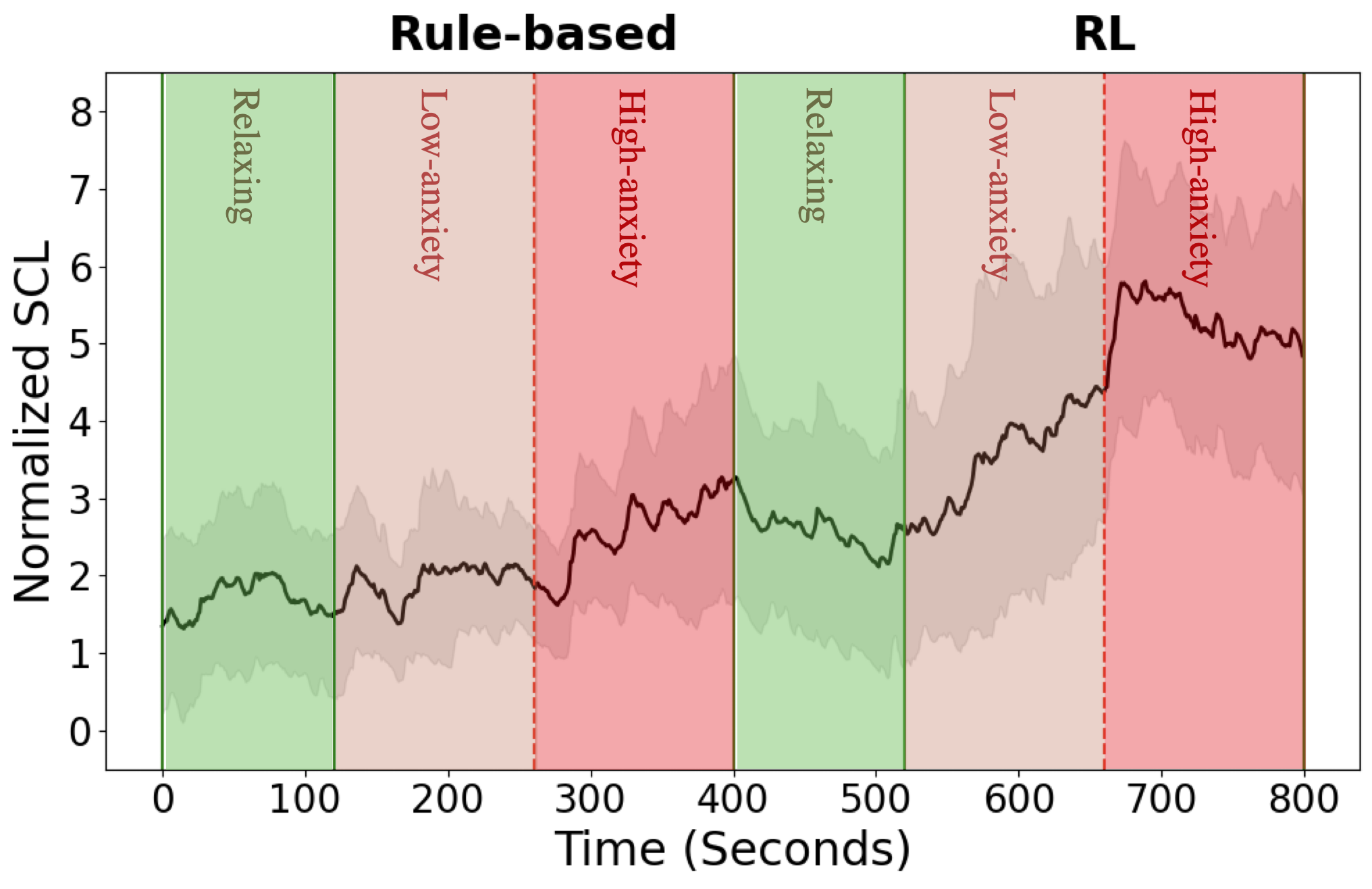}
  \end{minipage}
  \hfill
  \caption{Average SUDs (top) and SCL (bottom) of participants during the VR experience. The left plots display anxiety levels for participants first exposed to an anxious environment adapted by the RL method, followed by the rules-based method. The right plot depicts the opposite order. The shaded region represents the standard deviation. Participants were not asked to rate their SUDs during the relaxing environment.}

  \label{fig: SUDs_SCL}
\end{figure*}

\paragraph{SUDs Results:}
Figure~\ref{fig: SUDs_SCL} (top plots) showcase the SUDs data averaged over all participants.
In the analysis of self-reported anxiety levels, both the RL and rules-based methods effectively increased SUDs during high-anxiety segments. Specifically, nineteen participants (86\%) showed significantly higher SUDs ($p < 0.01$) using Wilcoxon signed-rank test in the high-anxiety segment when using the RL method, compared to sixteen participants (72\%) with the rules-based method. 
Additionally, when comparing the precision of achieving desired anxiety levels, the RL method exhibited a significantly lower Mean Squared Error (MSE) for high-anxiety levels (M=891.49) compared to the rules-based method (M=1538.12), indicating greater precision in reaching the target anxiety level.

\paragraph{EDA Sensor Results:}
Objective measurements using EDA provided further evidence of the framework's effectiveness.
Figure~\ref{fig: SUDs_SCL} (bottom plots) showcase the SCL data averaged over all participants.
The RL method consistently achieved higher SCL values during high-anxiety segments across all participants, whereas the rules-based method only did so for 72\% percent of participants. 
We employed Mean Squared Error (MSE) to evaluate accuracy. Both methods effectively reached the low-anxiety level, but the RL method was more precise and consistent overall. Importantly, the RL method exhibited significantly lower MSE (M=6.82) for the high-anxiety segment compared to the rules-based method (M=22.34) ($p < 0.01$), demonstrating greater accuracy in achieving the desired anxiety levels.

\begin{figure*}[t]
  \centering
  \begin{minipage}[b]{0.32\textwidth}
    \centering
    \includegraphics[width=\textwidth]{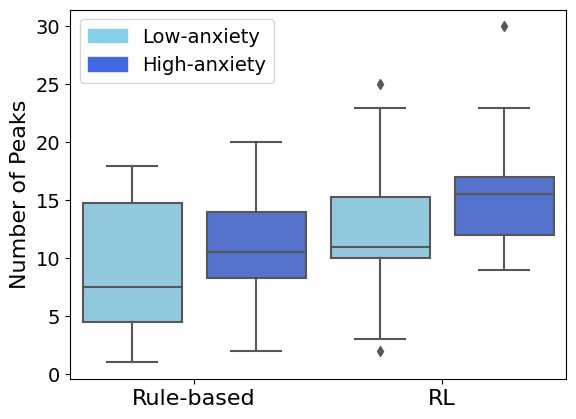}
  \end{minipage}
  \hfill
  \begin{minipage}[b]{0.32\textwidth}
    \centering
    \includegraphics[width=\textwidth]{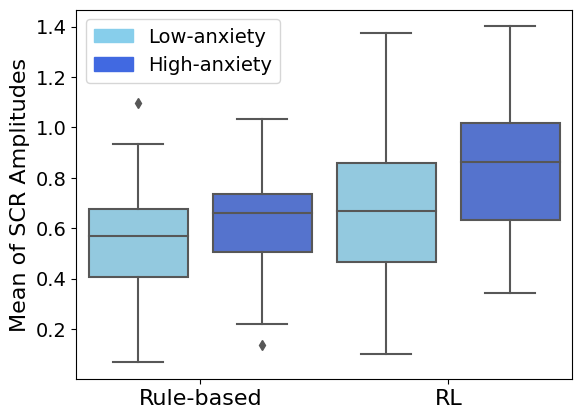}
  \end{minipage}
  \hfill
  \begin{minipage}[b]{0.32\textwidth}
    \centering
    \includegraphics[width=\textwidth]{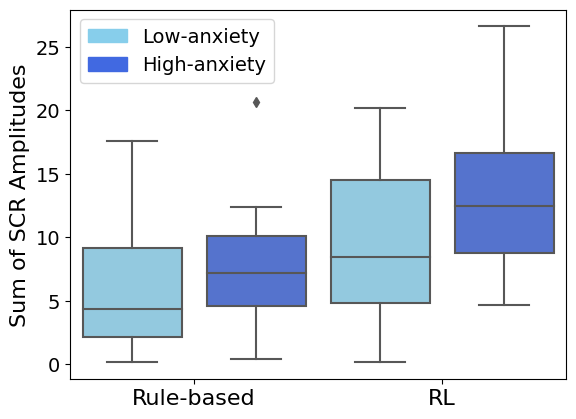}
  \end{minipage}
  \hfill
  \caption{Comparison of three SCR features between the low-anxiety and high-anxiety segments when the adaptive methods were set to the rules-based method or RL method. The leftmost plot gives the number of peaks, the middle plot gives the mean of SCR amplitudes, and the rightmost gives the sum of SCR amplitudes. Across all three features, the RL method has a significant difference between the low and high anxiety values.}

  \label{fig: SCR}
\end{figure*}

Prior studies have extracted the number of peaks and the mean or sum of SCR amplitudes to measure the intensity of anxiety~\cite{suso2019virtual, pick2018autonomic, albayrak2023fear}.
The RL method significantly ($p < 0.01$) increased the number of Skin SCR peaks and the mean and sum of SCR amplitudes during high-anxiety segments, whereas the rules-based method did not show significant changes in these measures. This result is shown in Figure~\ref{fig: SCR}.

\paragraph{Questionnaire Results:}
Post-study questionnaires indicated that participants perceived the RL-adapted environments as significantly ($p < 0.01$) more stressful compared to the rules-based method. Participants reported feeling more tension, fear, discomfort, and disgust in environments adapted by the RL method, whereas the rules-based method was perceived as more pleasant, relaxing, and boring. On the State-Trait Anxiety Inventory (STAI-6) scale, participants reported significantly higher anxiety levels (M=57.26) with the RL method compared to the rules-based method (M=46.2), using a one-tailed paired t-test.

\section{Spider Personalization Experiment}

Building on our demonstration of the EDPCGRL framework's ability to personalize virtual spiders to elicit specific anxiety responses, we identified a potential limitation: the concern that a single spider configuration might be sufficient to achieve the desired anxiety levels for all individuals. We hypothesized that different participants would respond more intensely to distinct spider attributes, aligning with prior arachnophobia research~\cite{lindner2019so}.

To test this hypothesis, we identified the spider that elicited the maximum anxiety response for each participant in our human subject study. We then employed K-Means clustering to group these spiders based on their attributes, identifying eight distinct clusters using the Elbow Method. Participants were associated with the cluster containing their most anxiety-inducing spider. The resulting distribution across clusters revealed significant diversity in spider attributes that triggered anxiety, indicating that a one-size-fits-all spider is insufficient for personalized VRET.

Further analysis assessed how participants reacted to spiders from different clusters. Participants predominantly exhibited heightened anxiety responses to spiders within their own cluster, while showing only mild responses to spiders from other clusters (see Table~\ref{Table:scares_clusters}). This pattern confirms that individual perceptions of anxiety-inducing spider features vary, thereby reinforcing the necessity for personalized spider adaptations in VRET. These findings underscore the importance of tailoring virtual stimuli to individual needs to enhance the effectiveness of exposure therapy for arachnophobia.

\begin{table}
\small
\begin{tabular}{|p{0.5cm}|*{9}{p{0.5cm}|}}
\hline
& C0 &  C1 & C2 & C3 & C4 & C5 & C6 & C7 \\
\hline
C0 & \heatmapcolor{7.5} & \heatmapcolor{3.5} & \heatmapcolor{4} & \heatmapcolor{4.5} & \heatmapcolor{4} & \heatmapcolor{4} & \heatmapcolor{4.5} & \heatmapcolor{4} \\ \hline
C1 & \heatmapcolor{3.5} & \heatmapcolor{8.25} & \heatmapcolor{5.66} & \heatmapcolor{3.5} & \heatmapcolor{6.5} & \heatmapcolor{3.33} & N/A & \heatmapcolor{3.75} \\ \hline
 C2 & \heatmapcolor{5} & \heatmapcolor{2.5} & \heatmapcolor{7.33} & \heatmapcolor{5} & \heatmapcolor{6} & \heatmapcolor{2.33} & \heatmapcolor{5} & \heatmapcolor{2.66} \\ \hline
 C3 & \heatmapcolor{2.5} & \heatmapcolor{5} & \heatmapcolor{5} & \heatmapcolor{8} & \heatmapcolor{3} & \heatmapcolor{7} & \heatmapcolor{4} & \heatmapcolor{3} \\ \hline
 C4 & \heatmapcolor{6} & \heatmapcolor{3} & \heatmapcolor{6} & \heatmapcolor{6} & \heatmapcolor{7.5} & \heatmapcolor{0} & \heatmapcolor{6.5} & \heatmapcolor{5.25} \\ \hline
 C5 & \heatmapcolor{2} & \heatmapcolor{6} & \heatmapcolor{3} & \heatmapcolor{7} & N/A & \heatmapcolor{9} & \heatmapcolor{5} & \heatmapcolor{7} \\ \hline
 C6 & \heatmapcolor{5.75} & \heatmapcolor{5.5} & \heatmapcolor{5.75} & \heatmapcolor{6} & \heatmapcolor{7.5} & \heatmapcolor{5.33} & \heatmapcolor{8.25} & \heatmapcolor{6.75} \\ \hline
 C7 & \heatmapcolor{5.5} & \heatmapcolor{3} & N/A & \heatmapcolor{6} & N/A & N/A & \heatmapcolor{6} & \heatmapcolor{7}      
\\
\hline

\end{tabular}
\caption{Maximum anxiety levels induced by spiders from different clusters, averaged across participants within each cluster. C\# represents the cluster number. ``N/A'' indicates that participants in that cluster never encountered spiders from the other cluster.}
\label{Table:scares_clusters}
\end{table}

\section{Discussion}

The results indicate that the RL-based EDPCGRL framework outperforms the rules-based method in both self-reported and physiological measures of anxiety. While both methods successfully increased anxiety levels, the RL method demonstrated greater consistency and precision, particularly in achieving higher anxiety levels. This suggests that the RL approach is more effective in personalizing the VRET experience, dynamically adjusting the virtual spider's attributes to match individual anxiety responses without relying on predefined rules.

Despite these promising results, 
we acknowledge several limitations. The study's reliance on a single physiological measure, Skin Conductance Level (SCL), may not capture the full complexity of anxiety responses. Future research should incorporate a broader range of physiological measures, such as EEG and gaze tracking, to provide a more comprehensive assessment of anxiety levels. 

Ethical considerations are crucial in the development and implementation of automated personalization in VRET. There is a potential risk of unintentionally causing distress or even re-traumatizing participants. To mitigate these risks, our system was designed to implement minimal changes and automatically terminate if a participant reached maximum anxiety levels. Future VR exposure therapy applications should be designed to avoid these potential negative outcomes.

Furthermore, its implementation within a controlled laboratory environment, while valuable for initial validation, does not indicate our approach is currently ready for broader clinical applications. Additionally, the relatively small sample size may not fully capture the diversity of responses necessary to validate the framework across different populations and anxiety disorders. To address these limitations, we intend to partner with therapists who employ arachnophobia exposure therapy. This would begin with a collaborative, human-centered design approach to ensure that our system meets their needs. Following this, we would seek clinical trials with patients suffering from arachnophobia to validate the effectiveness of our system in terms of patient outcomes. In these trials, we would seek a broad range of participants across various demographic groups to ensure generalizability. The last step would be to obtain certification through a local health services agency to support widespread adoption.

While our current method is based on a simple yet effective RL technique, we are interested in exploring more advanced RL techniques to assess whether their performance exhibits notable variations. Given the simplicity of our state space, we do not anticipate a significant difference in performance or user experience with a more complex RL technique. However, more complex environments may benefit from these techniques. 

Despite these limitations, the findings provide strong evidence for the potential of RL-based adaptive methods in enhancing the effectiveness of VRET for treating arachnophobia. The EDPCGRL framework represents a significant advancement over standard approaches for automating personalized exposure therapy, offering a more nuanced and responsive method for addressing individual anxiety responses. Future research should build on this foundation by incorporating additional physiological indicators, exploring long-term adaptability, and validating therapeutic outcomes through collaborative clinical studies. These efforts will be essential in fully realizing the benefits of reinforcement learning in personalized mental health interventions.

\section{Conclusion}

Enhancing the efficacy of Virtual Reality Exposure Therapy (VRET) often hinges on adapting to the unique needs of each individual instead of a one-size-fits-all approach.
In this work, we developed a novel framework for adaptive VRET using an Experience-Driven Procedural Content Generation via Reinforcement Learning
(EDPCGRL) approach.
We evaluated our framework via an implementation focusing on individuals with arachnophobia. We employed reinforcement learning in our implementation to modify a virtual spider in real-time based on a participant's anxiety indicators.
In our primary study, we evaluated the performance of our framework and compared it with a rules-based method.
Our results demonstrated that our framework can adapt virtual spiders to align with the desired anxiety levels in participants. 
Moreover, we observed a significant variance in user responses to different spider attributes, emphasizing the importance of a personalized therapeutic approach.
Returning to the motivating use case, we hope this framework can aid in providing personalized automated therapy that is more engaging and precise, but also less time-consuming for practitioners.

\section{Acknowledgements}
We are pleased to acknowledge that this paper has been accepted with revisions for publication in the full version of the ACM Transactions on Interactive Intelligent Systems.

\bibliography{aaai25}

\begin{thebibliography}{55}
\providecommand{\natexlab}[1]{#1}

\bibitem[{Albayrak et~al.(2023)Albayrak, Jablonski, Felderhoff-Mueser, Huening,
  Ernst, Timmann, and Batsikadze}]{albayrak2023fear}
Albayrak, B.; Jablonski, L.; Felderhoff-Mueser, U.; Huening, B.~M.; Ernst,
  T.~M.; Timmann, D.; and Batsikadze, G. 2023.
\newblock Fear conditioning is preserved in very preterm-born young adults
  despite increased anxiety levels.

\bibitem[{Alvares et~al.(2013)Alvares, Quintana, Kemp, Van~Zwieten, Balleine,
  Hickie, and Guastella}]{alvares2013reduced}
Alvares, G.~A.; Quintana, D.~S.; Kemp, A.~H.; Van~Zwieten, A.; Balleine, B.~W.;
  Hickie, I.~B.; and Guastella, A.~J. 2013.
\newblock Reduced heart rate variability in social anxiety disorder:
  associations with gender and symptom severity.
\newblock \emph{PloS one}, 8(7): e70468.

\bibitem[{Badia et~al.(2018)Badia, Quintero, Cameirao, Chirico, Triberti,
  Cipresso, and Gaggioli}]{i2018toward}
Badia, S.~B.; Quintero, L.~V.; Cameirao, M.~S.; Chirico, A.; Triberti, S.;
  Cipresso, P.; and Gaggioli, A. 2018.
\newblock Toward emotionally adaptive virtual reality for mental health
  applications.
\newblock \emph{IEEE journal of biomedical and health informatics}, 23(5):
  1877--1887.

\bibitem[{Balan et~al.(2020)Balan, Cristea, Moldoveanu, Moise, Leordeanu, and
  Moldoveanu}]{bualan2020towards}
Balan, O.; Cristea, S.; Moldoveanu, A.; Moise, G.; Leordeanu, M.; and
  Moldoveanu, F. 2020.
\newblock Towards a Human-Centered Approach for VRET Systems: case study for
  acrophobia.
\newblock In \emph{Advances in Information Systems Development: Information
  Systems Beyond 2020 28}, 182--197. Springer.

\bibitem[{Bian et~al.(2019)Bian, Wade, Swanson, Weitlauf, Warren, and
  Sarkar}]{bian2019design}
Bian, D.; Wade, J.; Swanson, A.; Weitlauf, A.; Warren, Z.; and Sarkar, N. 2019.
\newblock Design of a physiology-based adaptive virtual reality driving
  platform for individuals with ASD.
\newblock \emph{ACM Transactions on Accessible Computing (TACCESS)}, 12(1):
  1--24.

\bibitem[{Boucsein(2012)}]{boucsein2012electrodermal}
Boucsein, W. 2012.
\newblock \emph{Electrodermal activity}.
\newblock Springer Science \& Business Media.

\bibitem[{Bun et~al.(2017)Bun, Gorski, Grajewski, Wichniarek, and
  Zawadzki}]{bun2017low}
Bun, P.; Gorski, F.; Grajewski, D.; Wichniarek, R.; and Zawadzki, P. 2017.
\newblock Low--cost devices used in virtual reality exposure therapy.
\newblock \emph{Procedia Computer Science}, 104: 445--451.

\bibitem[{Carl et~al.(2019)Carl, Stein, Levihn-Coon, Pogue, Rothbaum,
  Emmelkamp, Asmundson, Carlbring, and Powers}]{carl2019virtual}
Carl, E.; Stein, A.~T.; Levihn-Coon, A.; Pogue, J.~R.; Rothbaum, B.; Emmelkamp,
  P.; Asmundson, G.~J.; Carlbring, P.; and Powers, M.~B. 2019.
\newblock Virtual reality exposure therapy for anxiety and related disorders: A
  meta-analysis of randomized controlled trials.
\newblock \emph{Journal of anxiety disorders}, 61: 27--36.

\bibitem[{Chalmers et~al.(2014)Chalmers, Quintana, Abbott, and
  Kemp}]{chalmers2014anxiety}
Chalmers, J.~A.; Quintana, D.~S.; Abbott, M. J.-A.; and Kemp, A.~H. 2014.
\newblock Anxiety disorders are associated with reduced heart rate variability:
  a meta-analysis.
\newblock \emph{Frontiers in psychiatry}, 5: 80.

\bibitem[{Cheng et~al.(2022)Cheng, Su, Liu, Huang, and Huang}]{cheng2022heart}
Cheng, Y.-C.; Su, M.-I.; Liu, C.-W.; Huang, Y.-C.; and Huang, W.-L. 2022.
\newblock Heart rate variability in patients with anxiety disorders: A
  systematic review and meta-analysis.
\newblock \emph{Psychiatry and Clinical Neurosciences}, 76(7): 292--302.

\bibitem[{Christian et~al.(2023)Christian, Cash, Cohen, Trombley, and
  Levinson}]{christian2023electrodermal}
Christian, C.; Cash, E.; Cohen, D.~A.; Trombley, C.~M.; and Levinson, C.~A.
  2023.
\newblock Electrodermal Activity and Heart Rate Variability During Exposure
  Fear Scripts Predict Trait-Level and Momentary Social Anxiety and
  Eating-Disorder Symptoms in an Analogue Sample.
\newblock \emph{Clinical Psychological Science}, 11(1): 134--148.

\bibitem[{Correa et~al.(2014)Correa, Cuervo, Perez, and Arias}]{correa2014new}
Correa, O.; Cuervo, C.; Perez, P.~C.; and Arias, A. 2014.
\newblock A new approach for self adaptive video game for
  rehabilitation-experiences in the amblyopia treatment.
\newblock In \emph{2014 IEEE 3nd International Conference on Serious Games and
  Applications for Health (SeGAH)}, 1--5. IEEE.

\bibitem[{C{\^o}t{\'e} and Bouchard(2009)}]{cote2009cognitive}
C{\^o}t{\'e}, S.; and Bouchard, S. 2009.
\newblock Cognitive mechanisms underlying virtual reality exposure.
\newblock \emph{CyberPsychology \& Behavior}, 12(2): 121--129.

\bibitem[{Dimovska et~al.(2010)Dimovska, Jarnfelt, Selvig, and
  Yannakakis}]{dimovska2010towards}
Dimovska, D.; Jarnfelt, P.; Selvig, S.; and Yannakakis, G.~N. 2010.
\newblock Towards procedural level generation for rehabilitation.
\newblock In \emph{Proceedings of the 2010 Workshop on Procedural Content
  Generation in Games}, 1--4.

\bibitem[{Dyrdal and Sanner(2022)}]{dyrdal2022virtual}
Dyrdal, O.; and Sanner, K. 2022.
\newblock \emph{Virtual Reality Exposure Therapy (VRET) For Fear of Spiders: a
  Randomized study}.
\newblock Master's thesis, NTNU.

\bibitem[{Fominykh et~al.(2018)Fominykh, Prasolova-F{\o}rland, Stiles, Krogh,
  and Linde}]{fominykh2018conceptual}
Fominykh, M.; Prasolova-F{\o}rland, E.; Stiles, T.~C.; Krogh, A.~B.; and Linde,
  M. 2018.
\newblock Conceptual framework for therapeutic training with biofeedback in
  virtual reality: First evaluation of a relaxation simulator.
\newblock \emph{Journal of Interactive Learning Research}, 29(1): 51--75.

\bibitem[{Gaebler et~al.(2013)Gaebler, Daniels, Lamke, Fydrich, and
  Walter}]{gaebler2013heart}
Gaebler, M.; Daniels, J.~K.; Lamke, J.-P.; Fydrich, T.; and Walter, H. 2013.
\newblock Heart rate variability and its neural correlates during emotional
  face processing in social anxiety disorder.
\newblock \emph{Biological psychology}, 94(2): 319--330.

\bibitem[{Gorini et~al.(2010)Gorini, Pallavicini, Algeri, Repetto, Gaggioli,
  and Riva}]{gorini2010virtual}
Gorini, A.; Pallavicini, F.; Algeri, D.; Repetto, C.; Gaggioli, A.; and Riva,
  G. 2010.
\newblock Virtual reality in the treatment of generalized anxiety disorders.
\newblock \emph{Annual Review of Cybertherapy and Telemedicine 2010}, 39--43.

\bibitem[{Guzdial, Snodgrass, and
  Summerville(2022{\natexlab{a}})}]{guzdial2022classical}
Guzdial, M.; Snodgrass, S.; and Summerville, A.~J. 2022{\natexlab{a}}.
\newblock Classical PCG.
\newblock In \emph{Procedural Content Generation via Machine Learning: An
  Overview}, 7--22. Springer.

\bibitem[{Guzdial, Snodgrass, and
  Summerville(2022{\natexlab{b}})}]{guzdial2022reinforcement}
Guzdial, M.; Snodgrass, S.; and Summerville, A.~J. 2022{\natexlab{b}}.
\newblock Reinforcement Learning PCG.
\newblock In \emph{Procedural Content Generation via Machine Learning: An
  Overview}, 159--179. Springer.

\bibitem[{Harrewijn et~al.(2018)Harrewijn, Van~der Molen, Verkuil, Sweijen,
  Houwing-Duistermaat, and Westenberg}]{harrewijn2018heart}
Harrewijn, A.; Van~der Molen, M.; Verkuil, B.; Sweijen, S.;
  Houwing-Duistermaat, J.; and Westenberg, P. 2018.
\newblock Heart rate variability as candidate endophenotype of social anxiety:
  A two-generation family study.
\newblock \emph{Journal of Affective Disorders}, 237: 47--55.

\bibitem[{Hocine et~al.(2015)Hocine, Goua{\"\i}ch, Cerri, Mottet, Froger, and
  Laffont}]{hocine2015adaptation}
Hocine, N.; Goua{\"\i}ch, A.; Cerri, S.~A.; Mottet, D.; Froger, J.; and
  Laffont, I. 2015.
\newblock Adaptation in serious games for upper-limb rehabilitation: an
  approach to improve training outcomes.
\newblock \emph{User Modeling and User-Adapted Interaction}, 25: 65--98.

\bibitem[{Kaczkurkin and Foa(2022)}]{kaczkurkin2022cognitive}
Kaczkurkin, A.~N.; and Foa, E.~B. 2022.
\newblock Cognitive-behavioral therapy for anxiety disorders: an update on the
  empirical evidence.
\newblock \emph{Dialogues in clinical neuroscience}.

\bibitem[{Kampmann et~al.(2016)Kampmann, Emmelkamp, Hartanto, Brinkman,
  Zijlstra, and Morina}]{kampmann2016exposure}
Kampmann, I.~L.; Emmelkamp, P.~M.; Hartanto, D.; Brinkman, W.-P.; Zijlstra,
  B.~J.; and Morina, N. 2016.
\newblock Exposure to virtual social interactions in the treatment of social
  anxiety disorder: A randomized controlled trial.
\newblock \emph{Behaviour research and therapy}, 77: 147--156.

\bibitem[{Khalifa et~al.(2020)Khalifa, Bontrager, Earle, and
  Togelius}]{khalifa2020pcgrl}
Khalifa, A.; Bontrager, P.; Earle, S.; and Togelius, J. 2020.
\newblock Pcgrl: Procedural content generation via reinforcement learning.
\newblock In \emph{Proceedings of the AAAI Conference on Artificial
  Intelligence and Interactive Digital Entertainment}, volume~16, 95--101.

\bibitem[{Kimani and Bickmore(2019)}]{kimani2019addressing}
Kimani, E.; and Bickmore, T. 2019.
\newblock Addressing public speaking anxiety in real-time using a virtual
  public speaking coach and physiological sensors.
\newblock In \emph{Proceedings of the 19th ACM International Conference on
  Intelligent Virtual Agents}, 260--263.

\bibitem[{Kritikos, Alevizopoulos, and
  Koutsouris(2021)}]{kritikos2021personalized}
Kritikos, J.; Alevizopoulos, G.; and Koutsouris, D. 2021.
\newblock Personalized virtual reality human-computer interaction for
  psychiatric and neurological illnesses: a dynamically adaptive virtual
  reality environment that changes according to real-time feedback from
  electrophysiological signal responses.
\newblock \emph{Frontiers in Human Neuroscience}, 15: 596980.

\bibitem[{Kritikos et~al.(2019)Kritikos, Tzannetos, Zoitaki, Poulopoulou, and
  Koutsouris}]{kritikos2019anxiety}
Kritikos, J.; Tzannetos, G.; Zoitaki, C.; Poulopoulou, S.; and Koutsouris, D.
  2019.
\newblock Anxiety detection from electrodermal activity sensor with movement \&
  interaction during virtual reality simulation.
\newblock In \emph{2019 9th International IEEE/EMBS Conference on Neural
  Engineering (NER)}, 571--576. IEEE.

\bibitem[{Lindner et~al.(2019)Lindner, Miloff, Reuterski{\"o}ld, Andersson, and
  Carlbring}]{lindner2019so}
Lindner, P.; Miloff, A.; Reuterski{\"o}ld, L.; Andersson, G.; and Carlbring, P.
  2019.
\newblock What is so frightening about spiders? Self-rated and self-disclosed
  impact of different characteristics and associations with phobia symptoms.
\newblock \emph{Scandinavian journal of psychology}, 60(1): 1--6.

\bibitem[{Lonsdorf et~al.(2017)Lonsdorf, Menz, Andreatta, Fullana, Golkar,
  Haaker, Heitland, Hermann, Kuhn, Kruse et~al.}]{lonsdorf2017don}
Lonsdorf, T.~B.; Menz, M.~M.; Andreatta, M.; Fullana, M.~A.; Golkar, A.;
  Haaker, J.; Heitland, I.; Hermann, A.; Kuhn, M.; Kruse, O.; et~al. 2017.
\newblock Don’t fear ‘fear conditioning’: Methodological considerations
  for the design and analysis of studies on human fear acquisition, extinction,
  and return of fear.
\newblock \emph{Neuroscience \& Biobehavioral Reviews}, 77: 247--285.

\bibitem[{Lyu and Bidarra(2023)}]{lyu2023procedural}
Lyu, S.; and Bidarra, R. 2023.
\newblock Procedural generation of challenges for personalized gait
  rehabilitation.
\newblock In \emph{Proceedings of the 18th International Conference on the
  Foundations of Digital Games}, 1--11.

\bibitem[{Maples-Keller et~al.(2017)Maples-Keller, Bunnell, Kim, and
  Rothbaum}]{maples2017use}
Maples-Keller, J.~L.; Bunnell, B.~E.; Kim, S.-J.; and Rothbaum, B.~O. 2017.
\newblock The use of virtual reality technology in the treatment of anxiety and
  other psychiatric disorders.
\newblock \emph{Harvard review of psychiatry}, 25(3): 103.

\bibitem[{Michaliszyn et~al.(2010)Michaliszyn, Marchand, Bouchard, Martel, and
  Poirier-Bisson}]{michaliszyn2010randomized}
Michaliszyn, D.; Marchand, A.; Bouchard, S.; Martel, M.-O.; and Poirier-Bisson,
  J. 2010.
\newblock A randomized, controlled clinical trial of in virtuo and in vivo
  exposure for spider phobia.
\newblock \emph{Cyberpsychology, Behavior, and Social Networking}, 13(6):
  689--695.

\bibitem[{Miloff et~al.(2019)Miloff, Lindner, Dafg{\aa}rd, Deak, Garke,
  Hamilton, Heinsoo, Kristoffersson, Rafi, Sindemark
  et~al.}]{miloff2019automated}
Miloff, A.; Lindner, P.; Dafg{\aa}rd, P.; Deak, S.; Garke, M.; Hamilton, W.;
  Heinsoo, J.; Kristoffersson, G.; Rafi, J.; Sindemark, K.; et~al. 2019.
\newblock Automated virtual reality exposure therapy for spider phobia vs.
  in-vivo one-session treatment: A randomized non-inferiority trial.
\newblock \emph{Behaviour research and therapy}, 118: 130--140.

\bibitem[{Minns et~al.(2019)Minns, Levihn-Coon, Carl, Smits, Miller, Howard,
  Papini, Quiroz, Lee-Furman, Telch et~al.}]{minns2019immersive}
Minns, S.; Levihn-Coon, A.; Carl, E.; Smits, J.~A.; Miller, W.; Howard, D.;
  Papini, S.; Quiroz, S.; Lee-Furman, E.; Telch, M.; et~al. 2019.
\newblock Immersive 3D exposure-based treatment for spider fear: A randomized
  controlled trial.
\newblock \emph{Journal of anxiety disorders}, 61: 37--44.

\bibitem[{M{\"u}hlberger et~al.(2008)M{\"u}hlberger, Sperber, Wieser, and
  Pauli}]{muhlberger2008virtual}
M{\"u}hlberger, A.; Sperber, M.; Wieser, M.~J.; and Pauli, P. 2008.
\newblock A virtual reality behavior avoidance test (VR-BAT) for the assessment
  of spider phobia.
\newblock \emph{Journal of CyberTherapy and Rehabilitation}, 1(2): 147--158.

\bibitem[{Nikoli{\'c} et~al.(2018)Nikoli{\'c}, Aktar, B{\"o}gels, Colonnesi,
  and de~Vente}]{nikolic2018bumping}
Nikoli{\'c}, M.; Aktar, E.; B{\"o}gels, S.; Colonnesi, C.; and de~Vente, W.
  2018.
\newblock Bumping heart and sweaty palms: Physiological hyperarousal as a risk
  factor for child social anxiety.
\newblock \emph{Journal of Child Psychology and Psychiatry}, 59(2): 119--128.

\bibitem[{{\"O}st et~al.(2023){\"O}st, Enebrink, Finnes, Ghaderi, Havnen,
  Kvale, Salomonsson, and Wergeland}]{ost2023cognitive}
{\"O}st, L.-G.; Enebrink, P.; Finnes, A.; Ghaderi, A.; Havnen, A.; Kvale, G.;
  Salomonsson, S.; and Wergeland, G.~J. 2023.
\newblock Cognitive behavior therapy for adult anxiety disorders in routine
  clinical care: A systematic review and meta-analysis.
\newblock \emph{Clinical Psychology: Science and Practice}.

\bibitem[{Owens and Beidel(2015)}]{owens2015can}
Owens, M.~E.; and Beidel, D.~C. 2015.
\newblock Can virtual reality effectively elicit distress associated with
  social anxiety disorder?
\newblock \emph{Journal of Psychopathology and Behavioral Assessment}, 37:
  296--305.

\bibitem[{Pick, Mellers, and Goldstein(2018)}]{pick2018autonomic}
Pick, S.; Mellers, J.~D.; and Goldstein, L.~H. 2018.
\newblock Autonomic and subjective responsivity to emotional images in people
  with dissociative seizures.
\newblock \emph{Journal of Neuropsychology}, 12(2): 341--355.

\bibitem[{Repetto et~al.(2013)Repetto, Gaggioli, Pallavicini, Cipresso,
  Raspelli, and Riva}]{repetto2013virtual}
Repetto, C.; Gaggioli, A.; Pallavicini, F.; Cipresso, P.; Raspelli, S.; and
  Riva, G. 2013.
\newblock Virtual reality and mobile phones in the treatment of generalized
  anxiety disorders: a phase-2 clinical trial.
\newblock \emph{Personal and Ubiquitous Computing}, 17: 253--260.

\bibitem[{{\v{S}}alkevicius et~al.(2019){\v{S}}alkevicius,
  Dama{\v{s}}evi{\v{c}}ius, Maskeliunas, and
  Laukien{\.e}}]{vsalkevicius2019anxiety}
{\v{S}}alkevicius, J.; Dama{\v{s}}evi{\v{c}}ius, R.; Maskeliunas, R.; and
  Laukien{\.e}, I. 2019.
\newblock Anxiety level recognition for virtual reality therapy system using
  physiological signals.
\newblock \emph{Electronics}, 8(9): 1039.

\bibitem[{Sevil et~al.(2017)Sevil, Hajizadeh, Samadi, Feng, Lazaro, Frantz, Yu,
  Brandt, Maloney, and Cinar}]{sevil2017social}
Sevil, M.; Hajizadeh, I.; Samadi, S.; Feng, J.; Lazaro, C.; Frantz, N.; Yu, X.;
  Brandt, R.; Maloney, Z.; and Cinar, A. 2017.
\newblock Social and competition stress detection with wristband physiological
  signals.
\newblock In \emph{2017 IEEE 14th International Conference on Wearable and
  Implantable Body Sensor Networks (BSN)}, 39--42. IEEE.

\bibitem[{Sevin{\c{c}}(2018)}]{sevincc2018language}
Sevin{\c{c}}, Y. 2018.
\newblock Language anxiety in the immigrant context: Sweaty palms?
\newblock \emph{International Journal of Bilingualism}, 22(6): 717--739.

\bibitem[{Smits, Powers, and Otto(2019)}]{smits2019personalized}
Smits, J.~A.; Powers, M.~B.; and Otto, M.~W. 2019.
\newblock \emph{Personalized exposure therapy: A person-centered
  transdiagnostic approach}.
\newblock Oxford University Press.

\bibitem[{Suso-Ribera et~al.(2019)Suso-Ribera, Fern{\'a}ndez-{\'A}lvarez,
  Garc{\'\i}a-Palacios, Hoffman, Bret{\'o}n-L{\'o}pez, Banos, Quero, and
  Botella}]{suso2019virtual}
Suso-Ribera, C.; Fern{\'a}ndez-{\'A}lvarez, J.; Garc{\'\i}a-Palacios, A.;
  Hoffman, H.~G.; Bret{\'o}n-L{\'o}pez, J.; Banos, R.~M.; Quero, S.; and
  Botella, C. 2019.
\newblock Virtual reality, augmented reality, and in vivo exposure therapy: a
  preliminary comparison of treatment efficacy in small animal phobia.
\newblock \emph{Cyberpsychology, Behavior, and Social Networking}, 22(1):
  31--38.

\bibitem[{Sutton and Barto(2018)}]{sutton2018reinforcement}
Sutton, R.~S.; and Barto, A.~G. 2018.
\newblock \emph{Reinforcement learning: An introduction}.
\newblock MIT press.

\bibitem[{Tu et~al.(2018)Tu, Bi, Hao, and Xing}]{tu2018breathcoach}
Tu, L.; Bi, C.; Hao, T.; and Xing, G. 2018.
\newblock BreathCoach: A smart in-home breathing training system with
  bio-feedback via VR game.
\newblock In \emph{Proceedings of the 2018 ACM International Joint Conference
  and 2018 International Symposium on Pervasive and Ubiquitous Computing and
  Wearable Computers}, 468--471.

\bibitem[{Valmaggia et~al.(2016)Valmaggia, Latif, Kempton, and
  Rus-Calafell}]{valmaggia2016virtual}
Valmaggia, L.~R.; Latif, L.; Kempton, M.~J.; and Rus-Calafell, M. 2016.
\newblock Virtual reality in the psychological treatment for mental health
  problems: An systematic review of recent evidence.
\newblock \emph{Psychiatry research}, 236: 189--195.

\bibitem[{Wilhelm et~al.(2005)Wilhelm, Pfaltz, Gross, Mauss, Kim, and
  Wiederhold}]{wilhelm2005mechanisms}
Wilhelm, F.~H.; Pfaltz, M.~C.; Gross, J.~J.; Mauss, I.~B.; Kim, S.~I.; and
  Wiederhold, B.~K. 2005.
\newblock Mechanisms of virtual reality exposure therapy: The role of the
  behavioral activation and behavioral inhibition systems.
\newblock \emph{Applied psychophysiology and biofeedback}, 30: 271--284.

\bibitem[{Wolpe(1990)}]{wolpe1990practice}
Wolpe, J. 1990.
\newblock \emph{The practice of behavior therapy}.
\newblock Pergamon press.

\bibitem[{Yang et~al.(2021)Yang, Fang, Chen, Zhang, Yin, Man, Yang, and
  Lu}]{yang2021global}
Yang, X.; Fang, Y.; Chen, H.; Zhang, T.; Yin, X.; Man, J.; Yang, L.; and Lu, M.
  2021.
\newblock Global, regional and national burden of anxiety disorders from 1990
  to 2019: results from the Global Burden of Disease Study 2019.
\newblock \emph{Epidemiology and psychiatric sciences}, 30: e36.

\bibitem[{Yannakakis and Togelius(2011)}]{yannakakis2011experience}
Yannakakis, G.~N.; and Togelius, J. 2011.
\newblock Experience-driven procedural content generation.
\newblock \emph{IEEE Transactions on Affective Computing}, 2(3): 147--161.

\bibitem[{Yee and Shiota(2015)}]{yee2015insecure}
Yee, C.~I.; and Shiota, M.~N. 2015.
\newblock An insecure base: Attachment style and orienting response to positive
  stimuli.
\newblock \emph{Psychophysiology}, 52(7): 905--909.

\bibitem[{Zahabi and Abdul~Razak(2020)}]{zahabi2020adaptive}
Zahabi, M.; and Abdul~Razak, A.~M. 2020.
\newblock Adaptive virtual reality-based training: a systematic literature
  review and framework.
\newblock \emph{Virtual Reality}, 24: 725--752.

\end{thebibliography}

\end{document}